\documentclass[sigconf]{acmart}

\usepackage{booktabs} 

\usepackage{amsbsy}
\usepackage{amsmath}
\usepackage{graphicx}
\usepackage{subfigure}
\usepackage{color}
\usepackage{booktabs}
\usepackage{bm}

\usepackage{natbib}
\usepackage{etoolbox}
\makeatletter
\patchcmd{\@makefntext}{\insertfootnotetext{#1}}{\insertfootnotetext{\scriptsize#1}}{}{}
\makeatother

\usepackage[normalem]{ulem} 

\newcommand{\vh}{{\mathbf{h}}}

\newcommand{\vz}{{\mathbf{z}}}

\newcommand{\cA}{{\mathcal{A}}}

\newcommand{\cF}{{\mathcal{F}}}

\newcommand{\cR}{{\mathcal{R}}}
\newcommand{\cS}{{\mathcal{S}}}

\newcommand{\grad}{{\nabla}}    

\newcommand{\bc}{\begin{center}}
\newcommand{\ec}{\end{center}}

\newcommand{\bdm}{\begin{displaymath}}
\newcommand{\edm}{\end{displaymath}}

\newcommand{\beq}{\begin{equation}}
\newcommand{\eeq}{\end{equation}}

\newcommand{\bfl}{\begin{flushleft}}
\newcommand{\efl}{\end{flushleft}}

\newcommand{\bt}{\begin{tabbing}}
\newcommand{\et}{\end{tabbing}}

\newcommand{\beqn}{\begin{eqnarray}}
\newcommand{\eeqn}{\end{eqnarray}}

\newcommand{\beqs}{\begin{align*}} 
\newcommand{\eeqs}{\end{align*}}  

\usepackage{comment}
\usepackage{algorithm,algorithmic}
\usepackage{hyperref}
\usepackage{enumitem}

\graphicspath{{./fig/}}

\newtheorem{assumption}{Assumption}

\setcopyright{rightsretained}

\acmDOI{10.475/123_4}

\acmISBN{123-4567-24-567/08/06}

\acmConference[WOODSTOCK'97]{ACM Woodstock conference}{July 1997}{El
  Paso, Texas USA} 
\acmYear{1997}
\copyrightyear{2016}

\acmPrice{15.00}

\begin{document}
\title{Collaborative Deep Reinforcement Learning}

\author{Kaixiang Lin}
\affiliation{%
  \institution{Computer Science and Engineering\\ Michigan State University}
  \streetaddress{428 S Shaw Ln.}
  \city{East Lansing} 
  \state{MI} 
  \postcode{48824}
}
\email{linkaixi@msu.edu}

\author{Shu Wang}
\affiliation{%
  \institution{Computer Science\\ Rutgers University}
  \streetaddress{57 US Highway 1}
  \city{New Brunswick} 
  \state{NJ} 
  \postcode{088901}
}
\email{sw498@cs.rutgers.edu}

\author{Jiayu Zhou}
\affiliation{%
  \institution{Computer Science and Engineering\\ Michigan State University}
  \streetaddress{428 S Shaw Ln.}
  \city{East Lansing} 
  \state{MI} 
  \postcode{48824}
}
\email{jiayuz@msu.edu}

\begin{abstract} 

Besides independent learning, human learning process is highly improved by 
summarizing what has been learned, communicating it with peers, 
and subsequently fusing knowledge from different sources to assist the
current learning goal. This \emph{collaborative learning} procedure ensures
that the knowledge is shared, continuously refined, and concluded from different perspectives
to construct a more profound understanding.
The idea of knowledge transfer has led to many advances in
machine learning and data mining, but significant challenges remain,
especially when it comes to reinforcement learning, heterogeneous model
structures, and different learning tasks. Motivated by human collaborative
learning, in this paper we propose a collaborative deep reinforcement learning
(CDRL) framework that performs adaptive knowledge transfer among heterogeneous
learning agents. 
Specifically, the proposed CDRL conducts a novel deep knowledge distillation method to
address the heterogeneity among different learning tasks with a deep alignment network.
Furthermore, we present an efficient
collaborative Asynchronous Advantage Actor-Critic (cA3C) algorithm to incorporate 
deep knowledge distillation into the online training of agents, and
demonstrate the effectiveness of the CDRL framework using extensive empirical
evaluation on OpenAI gym. 


\end{abstract}

%
%



\maketitle

\section{Introduction}

It is the development of cognitive abilities including learning, remembering,
communicating that enables human to conduct social cooperation, which is the
key to the rise of humankind. 
As a social animal, the ability to collaborate awoke the cognitive revolution and
reveals the prosperous history of human~\cite{Harari2015sapiens}.
In disciplines of cognitive science,
education and psychology, \emph{collaborative learning}, a situation in which
a group of people learn to achieve a set of tasks together, has been
advocated throughout previous studies~\cite{dillenbourg1999collaborative}. It
is intuitive to illustrate the concept of collaborative learning in the example
of group study. A group of students are studying together to master some
challenging course materials. As each student may understand the
materials from a distinctive perspective, effective communication would
greatly help the entire group achieve a better understanding than those from independent
study, and could significantly improve the efficiency and effectiveness of
learning process, as well~\cite{gokhale1995collaborative}.


\begin{figure}[t!]\vspace{5mm}
\centering
\includegraphics[width=0.48\textwidth]{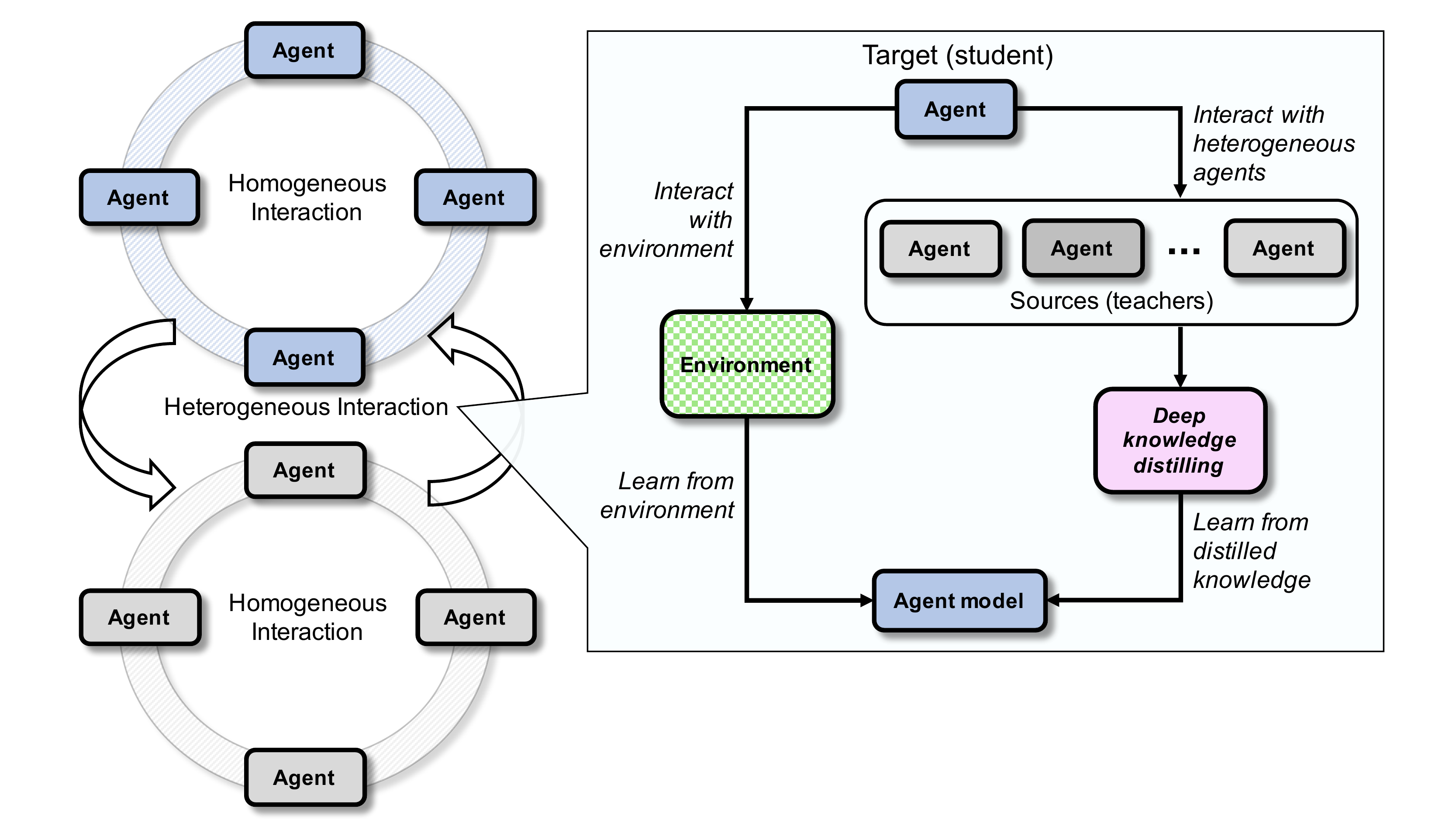}  
\caption{The illustration of Collaborative Deep Reinforcement Learning Framework.}
\label{fig:cdrl}\vspace{-3mm}
\end{figure}

On the other hand, the study of human learning has largely advanced the design
of machine learning and data mining algorithms, especially in reinforcement
learning and transfer learning. The recent success of deep reinforcement
learning (DRL) has attracted increasing attention from the community, as DRL
can discover very competitive strategies by having learning agents interacting
with a given environment and using rewards from the environment as the
supervision (e.g., \cite{mnih2015human,jaderberg2016reinforcement,
lample2016playing,silver2016mastering}). Even though most of current research
on DRL has focused on learning from games, it possesses great transformative
power to impact many industries with data mining and machine learning
techniques such as clinical decision support~\cite{thapa2005agent},
marketing~\cite{abe2004cross}, finance~\cite{abe2010optimizing}, visual
navigation~\cite{zhu2016target}, and autonomous
driving~\cite{carreras2005behavior}. Although there are many existing efforts
towards effective algorithms for
DRL~\cite{mnih2016asynchronous,nair2015massively}, the computational cost
still imposes significant challenges as training DRL for even a simple game
such as \textsc{Pong}~\cite{brockman2016openai} remains very expensive. The
underlying reasons for the obstacle of efficient training mainly lie in
two aspects: First, the supervision (rewards) from the environment is very
sparse and implicit during training. It may take an agent hundreds or even
thousands actions to get a single reward, and which actions that actually lead
to this reward are ambiguous. Besides the insufficient supervision, training
deep neural network itself takes lots of computational resources.

Due to the aforementioned difficulties, performing knowledge transfer
from other related tasks or well-trained deep models to facilitate training
has drawn lots of attention in the community~\cite{rajendran20152t,taylor2009transfer,
parisotto2015actor,jaderberg2016reinforcement,rusu2015policy}. 
Existing transfer learning can be categorized into two classes according to
the means that knowledge is transferred: {\it data
transfer}~\cite{hinton2015distilling,parisotto2015actor,rusu2015policy} and {\it model
transfer}~\cite{evgeniou2007multi,zhang2012convex,zhang2014facial,parisotto2015actor}. 
Model transfer methods implement knowledge transfer from introducing inductive
bias during the learning, and has been extensively studied in both transfer
learning/multi-task learning (MTL) community and deep learning community. For
example, in the regularized MTL models such as~\cite{evgeniou2004regularized,zhou2011malsar},
tasks with the same feature space are related through some structured
regularization. Another example is the multi-task deep neural network, where
different tasks share parts of the network structures~\cite{zhang2014facial}.
One obvious disadvantage of model transfer is the lack of flexibility: usually the feasibility
of inductive transfer has largely restricted the model structure of learning
task, which makes it not practical in DRL because for different tasks
the optimal model structures may be radically different.
On the other hand, the recently developed data transfer (also known as knowledge
distillation or mimic
learning)~\cite{hinton2015distilling,rusu2015policy,parisotto2015actor} embeds the 
source model knowledge into data points. Then they are used as knowledge bridge 
to train target models, which can have different structures as compared to
the source model~\cite{hinton2015distilling,bucilu2006model}. 
Because of the
structural flexibility, the data transfer is especially suitable to deal with
structure variant models. 


There are two situations that transfer learning methods are essential
in DRL:\\
\noindent \textbf{Certificated heterogeneous transfer.}
Training a DRL agent is computational expensive. If we have a 
well-trained model, it will be beneficial to assist the learning of other
tasks by transferring knowledge from this model. 
Therefore we consider following research question: Given one {\it
certificated} task (i.e. the model is well-designed, extensively trained and
performs very well), how can we maximize the information that can be used in
the training of other related tasks? Some model transfer approaches directly
use the weights from the trained model to initialize the new
task~\cite{parisotto2015actor}, which can only be done when the model
structures are the same. Thus, this strict requirement has largely limited
its general applicability on DRL. 
On the other hand, the initialization may not work
well if the tasks are significantly different from each other in nature~\cite{parisotto2015actor}. 
This challenge could be partially solved by generating an intermediate dataset (logits) from the 
existing model to help learning the new task.
However, new problems would arise when we are transferring knowledge between
{\it heterogeneous tasks}. Not only the action spaces are
different in dimension, the intrinsic action probability distributions and semantic
meanings of two tasks could differ a lot. 
Specifically, one action in \textsc{Pong} may refer
to move the paddle upwards while the same action index in \textsc{Riverraid}~\cite{brockman2016openai} would
correspond to fire. 
Therefore, the distilled dataset generated from the trained source task cannot 
be directly used to train the heterogeneous target task. In this scenario, the first key
challenge we identified in this work is that how to conduct data transfer among
heterogeneous tasks so that we can maximally utilize the information from a
certificated model while still maintain the flexibility of model design for
new tasks. During the transfer, the transferred knowledge from other tasks
may contradict to the knowledge that agents learned from its environment. One
recently work~\cite{rajendran20152t} use an attention network selective eliminate
transfer if the contradiction presents, which is not suitable in this setting
since we are given a certificated task to transfer. Hence, the second
challenge is how to resolve the conflict and perform a meaningful
transfer.


\noindent \textbf{ Lack of expertise.} 
A more general desired but also more challenging scenario is 
that DRL agents are trained for multiple heterogeneous tasks without 
any pre-trained models available.
One feasible way to conduct transfer under this scenario is
that agents of multiple tasks share part of their network
parameters~\cite{zhang2014facial,rusu2015policy}. 
However, an inevitable drawback is, 
multiple models lose their task-specific designs since 
the shared part needs to be the same. 
Another solution is to learn a domain invariant feature space
shared by all tasks~\cite{Abhishek2017learn}. However, some
task-specific information is often lost while converting the original state to a new feature
subspace. 
In this case, an intriguing questions is that: can we design a
framework that fully utilizes the original environment information and
meanwhile leverages the knowledge transferred from other tasks?

This paper investigates the aforementioned problems systematically and proposes a novel Collaborative Deep Reinforcement
Learning (CDRL) framework (illustrated in Figure~\ref{fig:cdrl}) to resolve them.
Our major contribution is threefold:
\begin{itemize}[leftmargin=0.2in]\itemsep0em 
\item First, in order to transfer knowledge among heterogeneous tasks while remaining the task-specific 
design of model structure, a novel deep knowledge distillation is proposed to address the heterogeneity 
among tasks, with the utilization of deep alignment network designed for the domain adaptation.
\item Second, in order to incorporate the transferred knowledge from heterogeneous tasks into the online 
training of current learning agents, similar to human collaborative learning, an efficient collaborative 
asynchronously advantage actor-critic learning (cA3C) algorithm is developed under the CDRL framework. 
In cA3C, the target agents are able to learn from environments and
its peers simultaneously, which also ensure the information from original environment is sufficiently utilized.
Further, the knowledge conflict among different tasks is resolved by adding an extra distillation layer to the policy network under CDRL framework, as well.
\item Last but not least we present extensive empirical studies on OpenAI gym to evaluate the proposed CDRL framework 
and demonstrate its effectiveness by achieving
more than 10\% performance improvement compared to the current state-of-the-art.
\end{itemize}

\textbf{Notations:}
In this paper, we use teacher network/source task denotes the network/task contained 
the knowledge to be transferred to others. Similarly, the student network/target 
task is referred to those tasks utilizing the knowledge transferred from 
others to facilitate its own training. The expert network denotes the network
that has already reached a relative high averaged reward in its own environment. In DRL, an agent is represented
by a policy network and a value network that share a set of parameters. Homogeneous agents
denotes agents that perform and learn under independent copies of same environment. Heterogeneous agents
refer to those agents that are trained in different environments.

\section{Related Work}

\noindent{\bf Multi-agent learning.}
One closely related area to our work is multi-agent reinforcement learning. A
multi-agent system includes a set of agents interacting in one environment.
Meanwhile they could potentially interact with each other~\cite{bucsoniu2010multi,kok2006collaborative,guestrin2002coordinated,tan1993multi}. In
collaborative multi-agent reinforcement learning, agents work together
to maximize a shared reward
measurement~\cite{kok2006collaborative,guestrin2002coordinated}. There is a
clear distinction between the proposed CDRL framework and multi-agent
reinforcement learning. In CDRL, each agent interacts with its own environment copy 
and the goal is to maximize the reward of the target agents. The formal definition of the proposed framework is given in
Section~\ref{sec:cdrlframework}.

\noindent{\bf Transfer learning.}
Another relevant research topic is domain adaption in the field of transfer
learning~\cite{pan2010survey,sun2011two,tzeng2015simultaneous}. The authors
in~\cite{sun2011two} proposed a two-stage domain adaptation framework that
considers the differences among marginal probability distributions of domains,
as well as conditional probability distributions of tasks.
The method first re-weights the data from the source domain using Maximum Mean Discrepancy and then re-weights the predictive function in the source domain to reduce 
te difference on conditional probabilities. In~\cite{tzeng2015simultaneous}, the marginal
distributions of the source and the target domain are aligned by training a network,
which maps inputs into a domain invariant representation. Also, knowledge
distillation was directly utilized to align the source and target class
distribution. One clear limitation here is that the source domain and
the target domain are required to have the same dimensionality (i.e. number of
classes) with same semantics meanings, which is not the case in our deep
knowledge distillation. 

In~\cite{Abhishek2017learn}, an invariant feature
space is learned to transfer skills between two agents. However, projecting the
state into a feature space would lose information contained in the original
state. There is a trade-off between learning the common feature space and
preserving the maximum information from the original state. In our work, we
use data generated by intermediate outputs in the knowledge
transfer instead of a shared space. Our approach thus retains complete
information from the environment and ensures high quality transfer. The recently
proposed A2T approach~\cite{rajendran20152t} can avoid negative
transfer among different tasks. However, it is possible that some negative
transfer cases may because of the inappropriate design of transfer algorithms. In
our work, we show that we can perform successful transfer among tasks that
seemingly cause negative transfer.
 
\noindent{\bf Knowledge transfer in deep learning.}
Since the training of each agent in an environment can be considered as a
learning task, and the knowledge transfer among multiple tasks belongs to the
study of multi-task learning. The multi-task deep neural network
(MTDNN)~\cite{zhang2014facial} transfers knowledge among tasks by sharing
parameters of several low-level layers. Since the low-level layers can be
considered to perform representation learning, the MTDNN is learning a shared
representation for inputs, which is then used by high-level layers in the
network. Different learning tasks are related to each other via this shared
feature representation. In the proposed CDRL, we do not use the share
representation due to the inevitable information loss when we project the
inputs into a shared representation. We instead perform explicitly knowledge
transfer among tasks by distilling knowledge that are independent of model
structures. In~\cite{hinton2015distilling}, the authors proposed to 
compress cumbersome models (teachers) to more simple models (students), where the simple models 
are trained by a dataset (knowledge) distilled from the teachers. However, this 
approach cannot handle the transfer among heterogeneous tasks, which is one key
challenge we addressed in this paper.

\noindent{\bf Knowledge transfer in deep reinforcement learning.}
Knowledge transfer is also studied in deep reinforcement learning. \cite{mnih2016asynchronous} proposed multi-threaded asynchronous variants of several most advanced deep reinforcement
learning methods including Sarsa, Q-learning, Q-learning and advantage 
actor-critic. Among all those methods, asynchronous advantage actor-critic (A3C)
achieves the best performance. Instead of using experience replay as in
previous work, A3C stabilizes the training procedure by training different
agents in parallel using different exploration strategies. This was shown to
converge much faster than previous methods and use less computational
resources. We show in Section~\ref{sec:cdrlframework} that the A3C is subsumed 
to the proposed CDRL as a special case. 
In~\cite{parisotto2015actor}, a single multi-task policy
network is trained by utilizing a set of expert Deep Q-Network (DQN) of source
games. At this stage, the goal is to obtain a policy network that can play
source games as close to experts as possible. The second step is to transfer
the knowledge from source tasks to a new but related target task. The
knowledge is transferred by using the DQN in last step as the
initialization of the DQN for the new task. As such, the training time of the new
task can be significantly reduced. Different from their approach, 
the proposed transfer strategy is not to directly mimic experts' actions or 
initialize by a pre-trained model.
In~\cite{rusu2015policy}, knowledge distillation was adopted to train a 
multi-task model that outperforms single task models of some tasks. 
The experts for all tasks are firstly acquired by single task learning. 
The intermediate outputs from each expert are then distilled to a similar multi-task
network with an extra controller layer to coordinate different action sets.
One clear limitation is that major components of the model are exactly the same
for different tasks, which may lead to degraded performance on some tasks. 
In our work, transfer can happen even when there are no experts available. Also,
our method allow each task to have their own model structures. 
Furthermore, even the model structures are the same for 
multiple tasks, the tasks are not trained to improve the performance of 
other tasks (i.e. it does not mimic experts from other tasks directly). Therefore our model
can focus on maximizing its own reward, instead of being distracted by others.

\section{Background}

\subsection{Reinforcement Learning}
In this work, we consider the standard reinforcement learning setting where 
each agent interacts with it's own environment over a number of discrete time 
steps.
Given the current state $s_t \in \cS$ at step $t$, agent $g_i$ selects 
an action $a_t \in \cA$ according to its policy 
$\pi(a_t|s_t)$, and receives a reward $r_{t+1}$ from the environment. 
The goal of the agent is to choose an action $a_t$ at step $t$ that 
maximize the sum of future rewards $\{r_t\}$ in a decaying manner:
$R_t = \sum_{i=0}^{\infty} \gamma^i r_{t+i}$, where scalar $\gamma \in (0, 1] $ is a discount rate.
Based on the policy $\pi$ of this agent, we can further define a state value function $V(s_t) = E{[R_t|s = s_t]}$, which estimates
the expected discounted return starting from state $s_t$, taking
actions following policy $\pi$ until the game ends.
%
The goal in reinforcement learning 
algorithm is to maximize the expected return. 
Since we are mainly discussing one specific agent's design and behavior throughout the paper, we 
leave out the notation of the agent index for conciseness.

\subsection{Asynchronous Advantage actor-critic algorithm (A3C)} 

The asynchronous advantage actor-critic (A3C) algorithm~\cite{mnih2016asynchronous} launches multiple
agents in parallel and asynchronously updates a global shared target policy network $\pi(a|s, \theta_p)$ as well as a value network $V(s,\theta_v)$.
parametrized by $\theta_p$ and $\theta_v$, respectively.
Each agent interacts with the environment, independently.
At each step $t$ the agent takes an action based on the probability distribution generated by
policy network. After playing a n-step rollout or reaching the terminal state, the rewards
are used to compute the advantage with the output of value function. The updates of policy network 
is conducted by applying the gradient:
$$\grad_{\theta_p} \log \pi(a_t| s_t; \theta_p)A(s_t, a_t;  \theta_v),$$
where the advantage function $A(s_t, a_t;\theta_v) $ is given by:
 $$\sum\nolimits_{i=0}^{T-t-1} \gamma^i r_{t+i} + \gamma^{T-t} V(s_{T};\theta_v) - V(s_t; \theta_v).$$
Term $T$ represents the step number for the last step of this rollout, it is either the max
number of rollout steps or the number of steps from $t$ to the terminal state. The update of value network
is to minimize the squared difference between the environment rewards and value function outputs, i.e., 
$$\min_{\theta_v} (\sum\nolimits_{i=0}^{T-t-1} \gamma^i r_{t+i} + \gamma^{T-t} V(s_{T};\theta_v) - V(s_t; \theta_v))^2.$$
The policy network and the value network share the same layers except for the last output layer.
An entropy regularization of policy $\pi$ is added to improve exploration, as well.

\subsection{Knowledge distillation}
Knowledge distillation~\cite{hinton2015distilling} is a transfer learning approach that
distills the knowledge from a teacher network to a student network using a temperature parameterized "soft targets" (i.e. a 
probability distribution over a set of classes). 
It has been shown that it can accelerate the training with less data since the gradient from "soft 
targets" contains much more information than the gradient obtained from "hard targets" (e.g. 0, 1 supervision).

To be more specific, logits vector $\vz \in \cR^d$ for $d$ actions can be converted to a probability distribution $\vh \in (0, 1)^d$ by a softmax function, raised with temperature $\tau$: 
\begin{align}
\vh(i) = \text{softmax}(\vz/\tau)_i = \frac{exp(\vz(i) / \tau)}{\sum_j exp(\vz(j) / \tau)}   
\label{eq:temperature_probability},
\end{align}
where $\vh(i)$ and $\vz(i)$ denotes the $i$-th entry of $\vh$ and $\vz$, respectively.

Then the knowledge distillation can be completed by optimize the following Kullback-Leibler divergence (KL)
with temperature $\tau$~\cite{rusu2015policy,hinton2015distilling}.
\begin{align}
L_{KL}(D, \theta_p^\beta) = \sum_{t=1} \text{softmax}(\vz_t^\alpha/\tau) \ln \frac{\text{softmax}(\vz_t^\alpha/\tau)}{\text{softmax}(\vz_t^\beta)}
\end{align}

where $\vz_t^\alpha$ is the logits vector from teacher network (notation $\alpha$ represents teacher) at step $t$, while $\vz_t^\beta$ is the logits 
vector from student network (notation $\beta$ represents student) of this step. 
$\theta_p^\beta$ denotes the parameters of the student policy network. $D$ is a set of logits from teacher network.

\section{Collaborative deep reinforcement learning framework}

In this section, we introduce the proposed collaborative deep reinforcement learning (CDRL) framework.
Under this framework, a collaborative Asynchronous Advantage Actor-Critic (cA3C) algorithm is proposed to 
confirm the effectiveness of the collaborative approach. 
Before we introduce our method in details, one underlying assumption we used
is as follows:
\begin{assumption}
If there is a universe that contains all the tasks $E = \{e_1, e_2, ..., e_\infty\}$ and $k_i$ represents the corresponding knowledge to master each task $e_i$, then $ \forall i, j, k_i \cap k_j \neq \emptyset $. 
\label{assumption:tasksimilarity}
\end{assumption}
This is a formal description of our common sense that any pair of
tasks are not absolutely isolated from each other, which has been implicitly used 
as a fundamental assumption by most prior transfer learning studies~\cite{parisotto2015actor,rusu2015policy,evgeniou2004regularized}.%
%
%
Therefore, we focus on  
mining the shared knowledge across multiple tasks instead of providing
strategy selecting tasks that share knowledge as much as possible, which remains
to be unsolved and may lead to our future work. The goal 
here is to utilize the existing knowledge as well as possible. For example,
we may only have a well-trained expert on playing Pong game, and we want to
utilize its expertise to help us perform better on 
other games. This is one of the situations that can be solved by our 
collaborative deep reinforcement learning framework.


\begin{figure*}[h!]
\centering
\begin{tabular}{c c}
\includegraphics[width=0.35\textwidth]{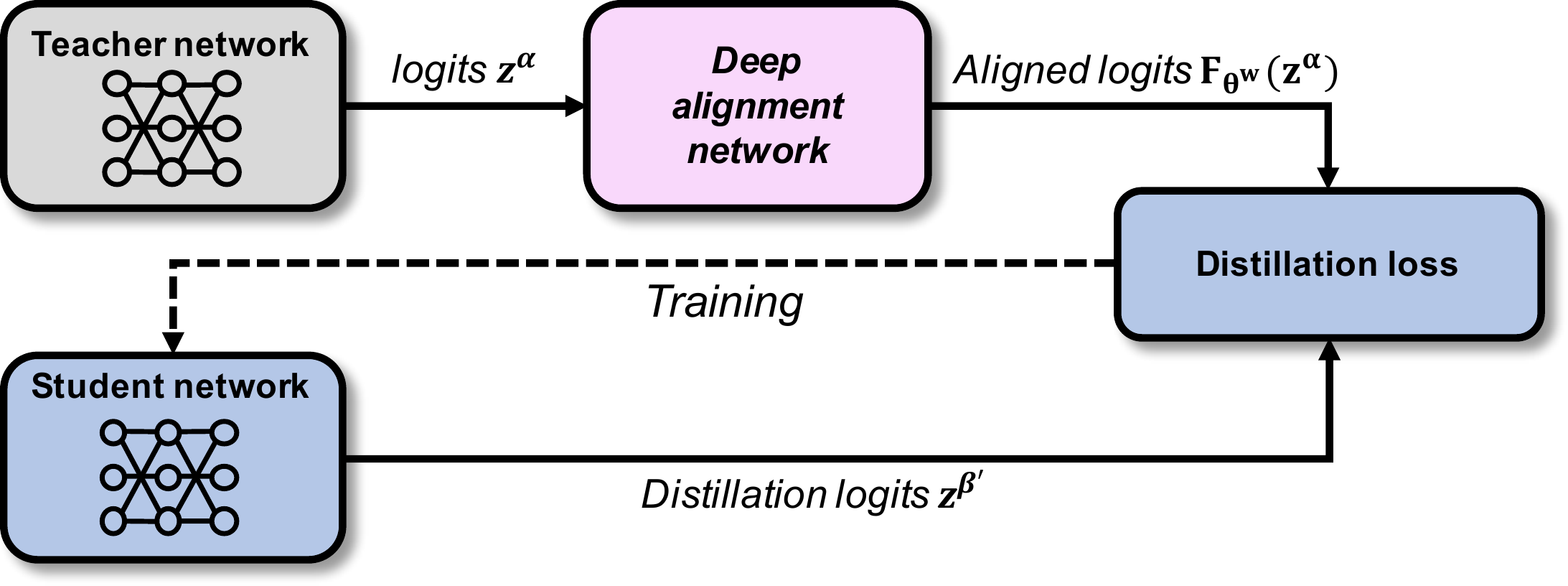} \quad\quad & 
\includegraphics[width=0.5\textwidth]{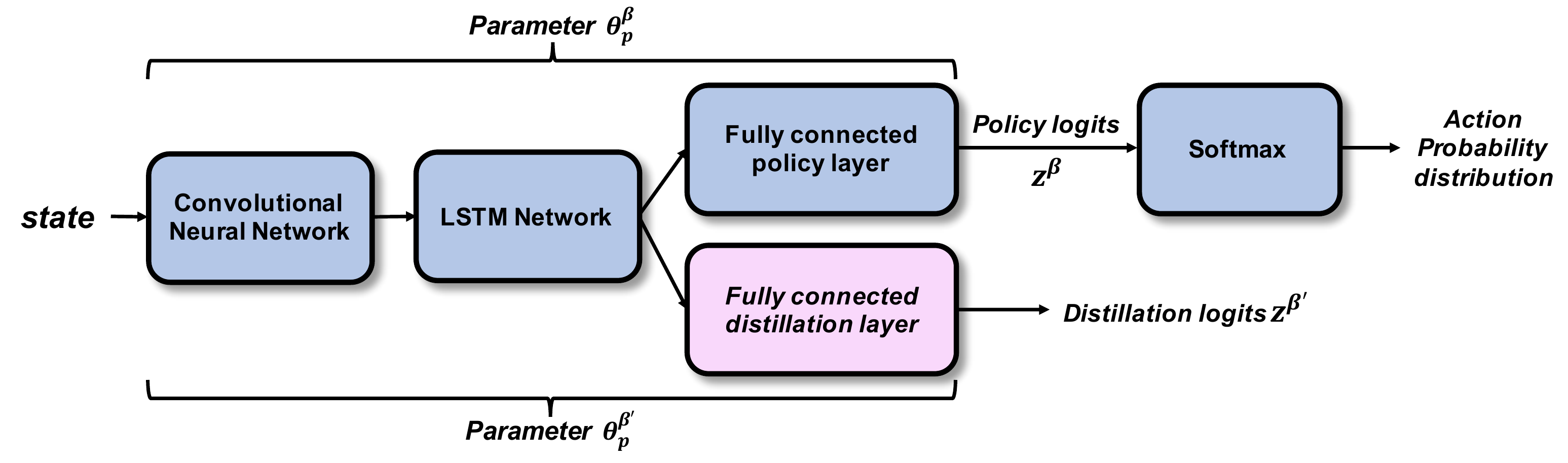}\\
(a) Distillation procedure & (b) Student network structure. \\
\end{tabular}
\vspace{-0.1in}
\caption{Deep knowledge distillation. In (a), the teacher's output logits $\vz^\alpha$ is mapped through a 
deep alignment network and the aligned logits $\cF_{\theta^\omega}(\vz^\alpha)$ is used as the supervision to train the student. .
In (b), the extra fully connected layer for distillation is added for learning knowledge from teacher. 
For simplicity's sake, time step $t$ is omitted here.}
\label{fig:deepKD}
\vspace{-0.1in}
\end{figure*}

\subsection{Collaborative deep reinforcement learning}\label{sec:cdrlframework}
In deep reinforcement learning, since the training of agents are computational 
expensive, the well-trained agents should be further utilized as source agents 
(agents where we transferred knowledge from) to facilitate the training of 
target agents (agents that are provided with the extra knowledge from source). 
In order to incorporate this type of collaboration to the training of DRL agents, 
we formally define the collaborative deep reinforcement learning (CDRL) framework as follows: 
\begin{definition}
Given $m$ independent environments $\{\varepsilon_1, \varepsilon_2, ..., \varepsilon_m\}$ of $m$ tasks
$\{e_1, e_2, ..., e_m\}$
, the corresponding $m$ agents
$\{g_1, g_2, ..., g_m\}$ are collaboratively trained
in parallel to maximize the rewards (master each task) with respect to target agents. 
\begin{itemize}[leftmargin=0.15in]\itemsep0em 
\item {\it Environments.}
There is no restriction on the environments: The $m$ environments can be totally different or with some duplications. 
\item {\it In parallel.}
Each environment $\varepsilon_i$ only interacts with the one corresponding agent $g_i$, i.e., 
the action $a_t^j$ from agent $g_j$ at step $t$ has no influence on the state $s_{t+1}^i$ in $\varepsilon_i, \forall i \neq j$.
\item {\it Collaboratively.}
The training procedure of agent $g_i$ consists of interacting with environment $\varepsilon_i$ and interacting
with other agents as well. The agent $g_i$ is not necessary to be at same level as "collaborative"
defined in cognitive science~\cite{dillenbourg1999collaborative}. E.g., $g_1$ can be an
expert for task $e_1$ (environment $\varepsilon_1$) while he is helping agent $g_2$ which is a student agent in task $e_2$.
\item {\it Target agents.}
The goal of CDRL can be set as maximizing the rewards that agent $g_i$ obtains in environment $\varepsilon_i$ 
with the help of interacting with other agents, similar to inductive transfer learning where $g_i$ is the target agent for target task
and others are source tasks. The knowledge is transfered from source to target $g_i$ by 
interaction.  
When we set the goal to maximize the rewards of multiple 
agents jointly, it is similar to multi-task learning where all tasks are source tasks and target
tasks at the same time.  
\end{itemize}
\end{definition}
Notice that our definition is very different from the previously defined collaborative 
multiagent Markow Decision Process (collaborative multiagent MDP)~\cite{kok2006collaborative,guestrin2002coordinated}
where a set of agents select a global joint action to maximize the sum of their individual 
rewards and the environment is transitted to a new state based on that joint action.
First, MDP is not a requirement in CDRL framework. Second, in CDRL, each agent has its own
copy of environment and maximizes its own cumulative rewards. The goal of collaboration
is to improve the performance of collaborative agents, compared with isolated ones, which
 is different from maximizing the sum of global rewards in collaborative multiagent MDP. 
Third, CDRL focuses on how agents collaborate among heterogeneous environments, instead of how joint action affects 
the rewards. In CDRL, different agents are acting in parallel, the actions taken by other agents
won't directly influence current agent's rewards. 
While in collaborative multiagent MDP, the agents must coordinate their action
choices since the rewards will be directly affected by the action choices of other agents. 

Furthermore, CDRL includes different types of interaction, which makes this a general framework.
For example, the current state-of-the-art is A3C~\cite{mnih2016asynchronous} can be categorized as one homogeneous CDRL method with advantage
actor-critic interaction. Specifically, multiple agents in A3C are trained in parallel with the same environment. All agents
first synchronize parameters from a global network, and then update the global network with their individual gradients.
This procedure can be seen as each agent maintains its own model (a different version of global network)
and interacts with other agents by sending and receiving gradients. 

In this paper, we propose a novel interaction method named deep knowledge distillation 
under the CDRL framework. 
It is worth noting that the interaction in A3C only deals with the homogeneous 
tasks, i.e. all agents have the same environment and the same model structure so that 
their gradients can be accumulated and interacted. 
By deep knowledge distillation, the interaction can be conducted among heterogeneous
tasks. 


\subsection{Deep knowledge distillation}
As we introduced before, knowledge distillation~\cite{hinton2015distilling} is 
trying to train a student network that can behave similarly to the teacher network
by utilizing the logits from the teacher as supervision. 
However, transferring the knowledge
among heterogeneous tasks faces several difficulties. 
First, the action spaces of different tasks may have different dimensions. 
Second, even if the dimensionality of
action space is same among tasks, the action probability distributions for different tasks
could vary a lot, as we illustrated in Figure~\ref{fig:action_prob} (a) and (b). Thus, the action
patterns represented by the logits of different policy networks are usually different from 
task to task.  If we directly 
force a student network to mimic the action pattern of a teacher network for a different task, it 
could be trained in a wrong direction, and finally ends up with worse performance than isolated training.
In fact, this suspect has been empirically verified in our experiments.

Based on the above observation, we propose deep knowledge distillation to transfer 
knowledge between heterogeneous tasks.
As illustrated in Figure~\ref{fig:deepKD}~(a), the approach for deep knowledge distillation is straightforward. 
We use a deep alignment network
to map the logits of the teacher network from a heterogeneous source task $e^\alpha$ (environment $\varepsilon^\alpha$), then the logits 
is used as our supervision to update the student network of target task $e^\beta$ (environment $\varepsilon^\beta$). This procedure is 
performed by minimizing following objective function over student policy network parameters ${\theta^\beta_p}'$:
\allowdisplaybreaks
\begin{align}
L_{KL}(D, {\theta^\beta_p}', \tau) &= \sum_{t} l_{KL}(\cF_{\theta^\omega}(\vz^\alpha_t), {\vz^\beta_t}', \tau), \; 
\label{eq:deepKDi}
\end{align}
where
\begin{align*}
l_{KL}(\cF_{\theta^\omega}(\vz^\alpha_t), {\vz^\beta_t}', \tau)
&= \text{softmax}(\cF_{\theta^\omega}(\vz^\alpha_t)/\tau) \ln \frac{\text{softmax}(\cF_{\theta^\omega}(\vz^\alpha_t)/\tau)}{\text{softmax}({\vz^\beta_t}')}\nonumber.
\end{align*}
Here $\theta^\omega$ denotes the parameters of the deep alignment network, which transfers the logits $\vz_t^{\alpha}$ from the
teacher policy network
for knowledge distillation by function $\cF_{\theta^\omega}(\vz^{\alpha}_t)$ at step $t$.
As we show in Figure~\ref{fig:deepKD} (b), $\theta_p^{\beta}$ is the student policy network parameters (including parameters of CNN, LSTM and policy layer) for task $e^{\beta}$, while ${\theta_p^{\beta}}'$ denotes student network parameters of CNN, LSTM and distillation layer. It is clear that
the distillation logits ${\vz^{\beta}_t}'$ from the student network does not determine the action probability 
distribution directly, 
which is established by the policy logits $\vz^{\beta}_t$, as illustrated in Figure~\ref{fig:deepKD} (b). 
We add another fully connected distillation layer to deal with the mismatch of action space dimensionality
and the contradiction of the transferred knowledge from source domain and the learned knowledge from target domain.
The input to both of the teacher and the student network is the state of environment $\varepsilon^\beta$ of target task $e^{\beta}$. It means
that we want to transfer the expertise from
an expert of task $e^{\alpha}$ towards the current state. 
Symbol $D$ is a set of logits from the teacher network in one batch and $\tau$ is the temperature same as described in Eq~(\ref{eq:temperature_probability}).
In a trivial case that the teacher network and the student network are trained for same task ($e^{\alpha}$ equals $e^{\beta}$), then the deep alignment network $\cF_{\theta^\omega}$ would 
reduce to an identity mapping, and the problem is also reduced to a single task policy distillation, which 
has been proved to be effective in~\cite{rusu2015policy}. 
Before we can apply the deep knowledge distillation, 
 we need to first train a good deep alignment network. In this work,
we provide two types of training protocols for different situations:\\
\noindent \textbf{Offline training}: 
This protocol first trains two teacher networks in both environment $\varepsilon^{\alpha}$ and $\varepsilon^{\beta}$. Then we use the logits
of both two teacher networks to train a deep alignment network $\cF_{\theta^\omega}$. After acquiring a pre-trained
$\cF_{\theta^\omega}$, we train a student network of task $e^{\beta}$ from scratch, in the meanwhile the teacher 
network of task $e^{\alpha}$ and $\cF_{\theta^\omega}$ are used for deep knowledge distillation. \\
\noindent \textbf{Online training}: Suppose we only have a teacher network of task $e^{\alpha}$, and we 
want to use the knowledge from task $e^{\alpha}$ to train the student network for task $e^{\beta}$ to get
higher performance from scratch. The pipeline of this method is that, we firstly 
train the student network by interacting with the environment $\varepsilon^\beta$ for a certain amount of 
steps $T_1$, 
and then start to train the alignment network $\cF_{\theta^\omega}$, 
using the logits from the teacher network and the student network. 
Afterwards, at step $T_2$, we start performing deep knowledge distillation. Obviously $T_2$
is larger than $T_1$, and the value of them are task-specific, which is 
decided empirically in this work.  

The offline training could be useful if we have already had a reasonably good model for
task $e^{\beta}$, while  we want to further improve the performance using the
knowledge from task $e^{\alpha}$. The online training method is used when we need to
learn the student network from scratch. Both types of training protocol can be 
extended to multiple heterogeneous tasks. 

\subsection{Collaborative Asynchronous Advantage\\ Actor-Critic}
In this section, we introduce the proposed collaborative asynchronous
advantage actor-critic (cA3C) algorithm. As we described in section~\ref{sec:cdrlframework}, 
the agents are running in parallel. Each agent goes through the same training
procedure as described in Algorithm~\ref{alg:mtlA3C}. 
As it shows, the training of 
agent $g_1$ can be separated into two parts: The first part is to interact with
the environment, get the reward and compute the gradients to minimize the value loss
and policy loss based on Generalized Advantage Estimation (GAE)~\cite{schulman2015high}.  The second part is to 
interact with source agent $g_2$ so that the logits distilled from agent $g_2$
can be transferred by the deep alignment network and used as supervision
to bias the training of agent $g_1$. 

To be more concrete, the pseudo code in Algorithm~\ref{alg:mtlA3C} is an envolved version 
of A3C based on online training of deep knowledge distillation. At $T$-th iteration, the 
agent interacts with the environment for $t_{max}$ steps or until the terminal state is 
reached (Line 6 to Line 15). Then the updating of value network
and policy network is conducted by GAE. This variation
of A3C is firstly implemented in OpenAI universe starter agent~\cite{OpenAIuniversestarteragent}.
Since the main asynchronous framework is the same as
A3C, we still use the A3C to denote this algorithm although the updating is the not the same as advantage actor-critic 
algorithm used in original A3C paper~\cite{mnih2016asynchronous}. 

The online training of deep knowledge distillation is mainly completed from Line 25 to Line 32 in Algorithm~\ref{alg:mtlA3C}. 
The training of the deep alignment network starts from $T_1$ steps (Line 25 - 28). 
After $T_1$ steps, the student network is able to generate a representative action probability 
distribution, and we have suitable supervision 
to train the deep alignment network as well, parameterized by $\theta^\omega$. 
After $T_2$ steps, $\theta^\omega$ will gradually converge to a local optimal, and we start 
the deep knowledge distillation. 
%
As illustrated in Figure~\ref{fig:deepKD}~(b), we use symbol ${\theta_p^\beta}'$ to represent the parameters
of CNN, LSTM and the fully connected distillation layer, since we don't want the logits from heterogeneous
directly affect the action pattern of target task. 
To simplify the discussion, the above algorithm is described based on interacting with a single agent from a 
heterogeneous task. 
In algorithm~\ref{alg:mtlA3C}, logits $\vz^{\alpha}_t$
can be acquired from multiple teacher networks of different tasks, each task will train its own
deep alignment network $\theta^\omega$ and distill the aligned logits to the student network. 

As we described in previous section~\ref{sec:cdrlframework}, there are two types of interactions in 
this algorithm: 1). GAE interaction uses the gradients shared by all homogeneous agents. 2) Distillation interaction 
is the deep knowledge distillation from teacher network. The GAE interaction is performed only among 
homogeneous tasks. By synchronizing the parameters from a global student network in Algorithm~\ref{alg:mtlA3C} (line 3), 
the current agent receives the GAE updates from all the other agents who interactes with the same environment.
In line 21 and 22, the current agent sends his gradients to the global student network, which will be synchronized
with other homogeneous agents. The distillation interaction is then conducted in line 31, where we have the aligned 
logits $\cF_{\theta^\omega}(\vz_t^\alpha)$ and the distillation logits ${\vz_t^\beta}'$ to compute the gradients for 
minimizing the distillation loss. 
The gradients of distillation are also sent to the global student network. 
The role of global student network can be regarded as a parameter server that helps sending interactions among 
the homogeneous agents. From a different angle, each homogeneous agent maintains an instinct version of global student network.
Therefore, both two types of interactions affect all homogeneous agents, which means that the distillation interactions 
from agent $g_2$ and agent $g_1$ would affect all homogeneous agents of agent $g_1$\footnote{Code is publicly available at \url{https://github.com/illidanlab/cdrl}}. 
\begin{algorithm}[h!]\small
\begin{algorithmic}[1]
\REQUIRE Global shared parameter vectors $\Theta_p$ and $\Theta_v$ and global shared counter $T = 0$;
Agent-specific parameter vectors $\Theta_p'$ and $\Theta_v'$, GAE~\cite{schulman2015high} parameters
$\gamma$ and $\lambda$. Time step to start training deep alignment network and deep knowledge distillation $T_1, T_2$.
\WHILE { $T<T_{max}$}
\STATE Reset gradients: $d\theta_p = 0$ and $d\theta_v = 0$
\STATE Synchronize agent-specific parameters $\theta_p' = \theta_p$ and $\theta_v' = \theta_v$
\STATE $t_{start} = t$, Get state $s_t$
\STATE Receive reward $r_t$ and new state $s_{t+1}$
\REPEAT
\STATE Perform $a_t$ according to policy 
\STATE Receive reward $r_t$ and new state $s_{t+1}$
\STATE Compute value of state $v_t =  V(s_t; \theta_v')$
\IF {$T \geq T_1$ }
\STATE Compute the logits $\vz_t^\alpha$ from teacher network. 
\STATE Compute the policy logits $\vz_t^\beta$ and distillation logits ${\vz_t^\beta}'$ from student network. 
\ENDIF
\STATE $t= t +1, T= T+1$
\UNTIL{terminal $s_t$ or $t- t_{start} >=t_{max}$} 
\STATE {\[ 
R = v_{t} =\begin{cases}
               0  \hspace{4.3em} \text{ for terminal } s_t \\
               V(s_t, \theta_{v}') \hspace{1em} \text{ for non-terminal } s_t  \\
            \end{cases}
\]}
\FOR {$ i \in \{t-1, ..., t_{start}\}$}
\STATE $\delta_i = r_i + \gamma v_{i+1} - v_{i}$
\STATE $ A = \delta_i + (\gamma\lambda)A$
\STATE $ R = r_i + \gamma R$
\STATE  $ d\theta_p \leftarrow d\theta_p + \nabla \log \pi(a_i | s_i; \theta')A$ 
\STATE  $ d\theta_v \leftarrow d\theta_v + \partial (R - v_i)^2/\partial \theta_v'$ 
\ENDFOR
\STATE Perform asynchronous update of $\theta_p$ using $d\theta_p$ and of $\theta_v$ using $d\theta_v$.
\IF {$T \geq T_1$ }
\STATE //  Training deep alignment network.
\STATE $\min_{\theta^\omega} \sum_{t} l_{KL}(\vz_t^\beta, \vz_t^\alpha, \tau)$, $l_{KL}$ is defined in Eq~(\ref{eq:deepKDi}).
\ENDIF
\IF {$T \geq T_2$ }
\STATE // online deep knowledge distillation.
\STATE $\min_{{\theta_p^\beta}'}  \sum_{t}  l_{KL}(\cF_{\theta^\omega}(\vz_t^\alpha), {\vz_t^\beta}')$
\ENDIF 
\ENDWHILE
\end{algorithmic}
\caption{online cA3C}
\label{alg:mtlA3C}
\end{algorithm}

\label{proof} 

\vspace{-0.2in}
\section{Experiments}
\subsection{Training and Evaluation}
In this work, training and evaluation are conducted in OpenAI
Gym~\cite{brockman2016openai}, a toolkit that includes a collection of
benchmark problems such as classic Atari games using Arcade Learning
Environment (ALE)~\cite{bellemare2013arcade}, classic control games, etc. Same
as the standard RL setting, an agent is stimulated in an
environment, taking an action and receiving rewards and observations at each time
step. The training of the agent is divided into episodes, and the goal is to maximize
the expectation of the total reward per episode or to reach higher performance
using as few episodes as possible.
\vspace{-0.12in}

\subsection{Certificated Homogeneous transfer}

In this subsection, we verify the effectiveness of knowledge distillation as a
type of interaction in collaborative deep reinforcement learning for
homogeneous tasks. This is also to verify the effectiveness of the simplest
case for deep knowledge distillation. Although the effectiveness of policy
distillation in deep reinforcement learning has been verified
in~\cite{rusu2015policy} based on DQN, there is no prior studies on
asynchronous online distillation. Therefore, our first experiment is to
demonstrate that the knowledge distilled from a certificated task can be used
to train a decent student network for a homogeneous task. Otherwise, the even
more challenging task of transferring among heterogeneous sources may not
work. We note that in this case, the
Assumption~\ref{assumption:tasksimilarity} is fully satisfied given $k_1 =
k_2$, where $k_1$ and $k_2$ are the knowledge needed to master task $e_1$ and
$e_2$, respectively. 
In this experiment, we conduct experiments in a gym environment named \textsc{Pong}.
It is a classic Atari game that an agent controls a paddle to bounce a ball pass
another player agent. 
The maximum reward that each episode can reach is 21. 

\begin{figure}
\centering
\begin{tabular}{c c}
\includegraphics[width=0.22\textwidth]{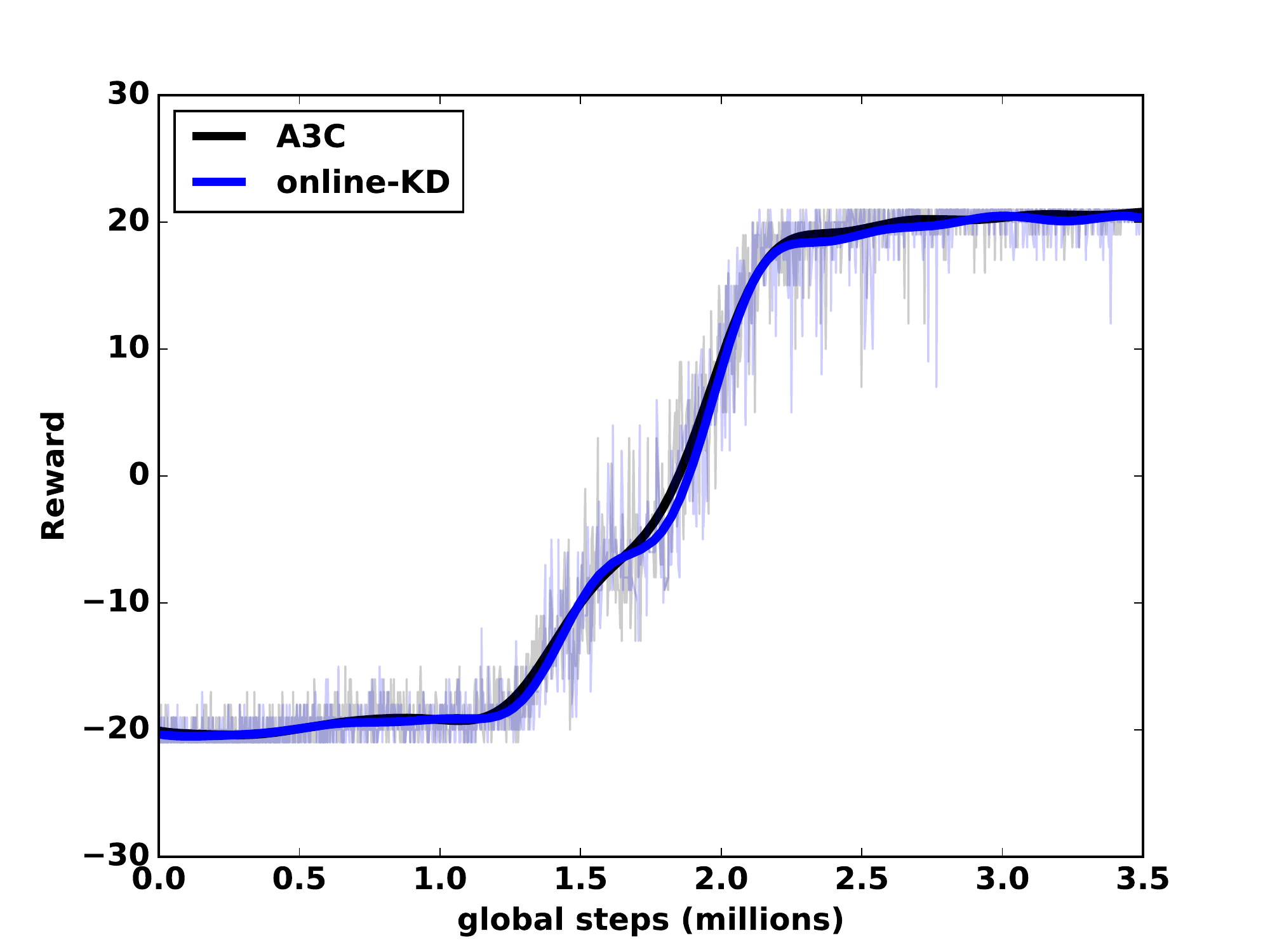} &   
\includegraphics[width=0.22\textwidth]{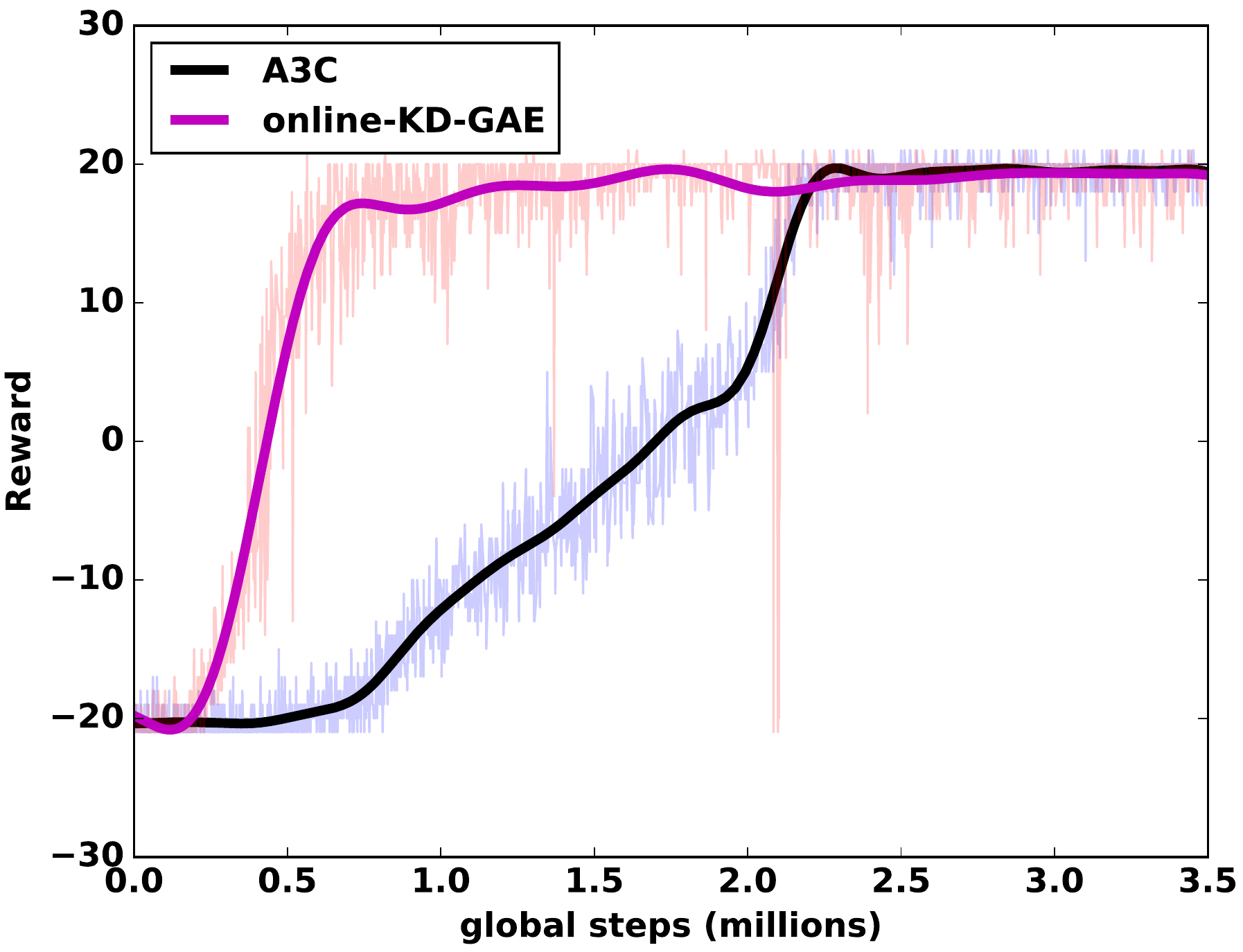} \\
(a) online KD only  & (b) online KD with GAE  \\
\end{tabular}
\vspace{-0.15in}
\caption{Performance of online homogeneous knowledge distillation. 
The results show that the combination of
knowledge distillation and GAE is an effective training strategy for
homogeneous tasks.
}
\label{fig:mtl-agent-distill}
\vspace{-0.2in}
\end{figure}

First, we train a teacher network that learns from its own environment by
asynchronously performing GAE updates. We then train a student network using
only online knowledge distillation from the teacher network. For fair
comparisons, we use 8 agents for all environments in the experiments.
Specifically, both the student and the teacher are training in
\textsc{Pong} with 8 agents. The 8 agents of the teacher network are trained using
the A3C algorithm (equivalent to CDRL with GAE updates in one task). The 8
agents of student network are trained using normal policy distillation, which
uses the logits generated from the teacher network as supervision to train the
policy network directly. From the results in Figure~\ref{fig:mtl-agent-distill} (a) we see that the
student network can achieve a very competitive performance that is is almost
same as the state-of-arts, using online knowledge distillation from a
homogeneous task. It also suggests that the teacher doesn't necessarily need
to be an expert, before it can guide the training of a student in the
homogeneous case. Before 2 million steps, the teacher itself is still learning
from the environment, while the knowledge distilled from teacher can already
be used to train a reasonable student network. Moreover, we see that the
hybrid of two types of interactions in CDRL has a positive effect on the
training, instead of causing performance deterioration.

In the second experiment, the student network is learning from both the online
knowledge distillation and the GAE updates from the environment. We find that
the convergence is much faster than the state-of-art, as shown in
Figure~\ref{fig:mtl-agent-distill} (b). In this experiment, the knowledge is
distilled from the teacher to student in the first one million steps and the
distillation is stopped after that. We note that in homogeneous CDRL,
knowledge distillation is used directly with policy logits other than
distillation logits. The knowledge transfer setting in this experiment is not
a practical one because we already have a well-trained model of \textsc{Pong},
but it shows that when knowledge is correctly transferred, the combination of
online knowledge distillation and the GAE updates is an effective training
procedure. 

\vspace{-0.12in}
\subsection{Certificated Heterogeneous Transfer}
\vspace{-0.05in}

\begin{figure}
\centering
\begin{tabular}{c c}
\includegraphics[width=0.22\textwidth]{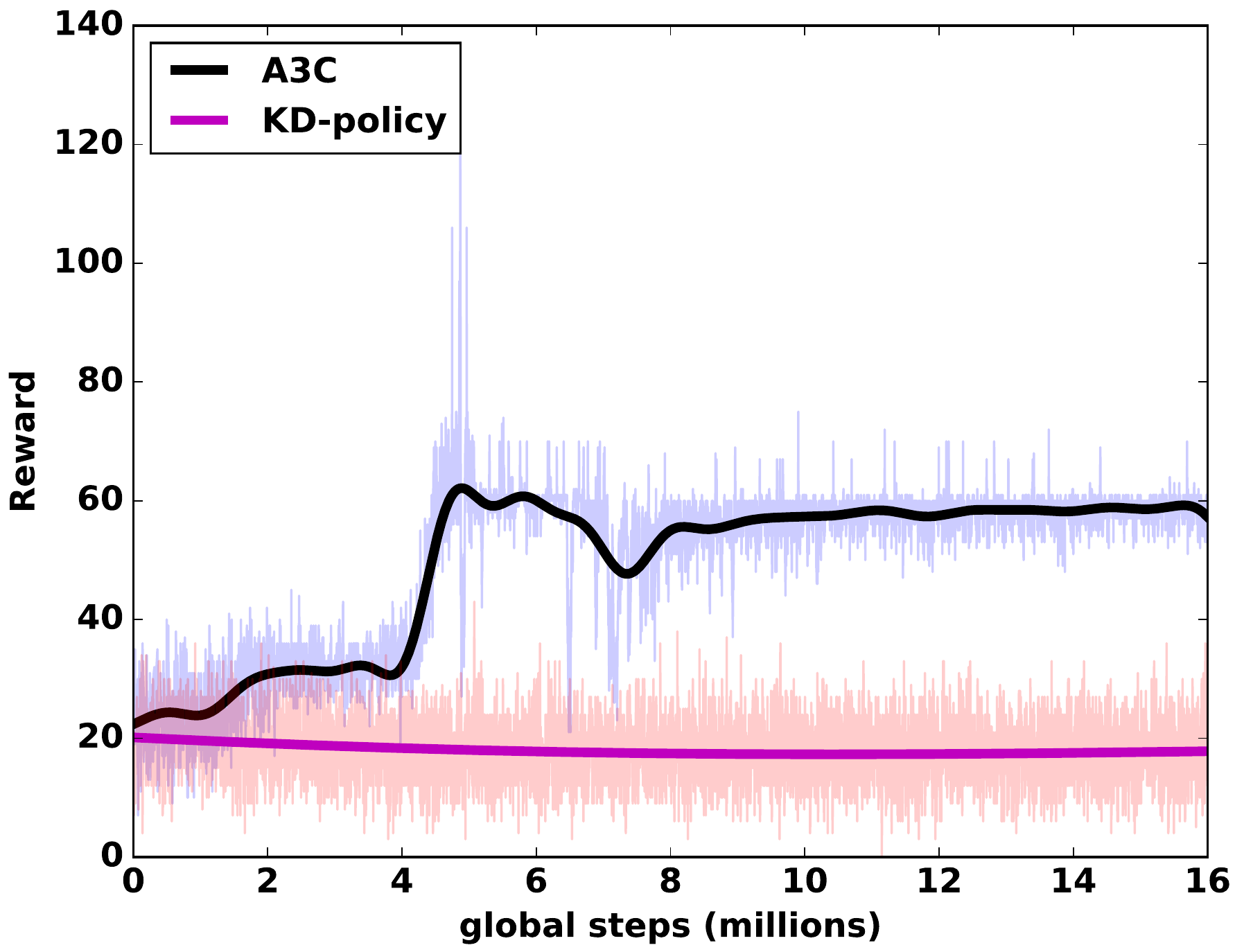} &
\includegraphics[width=0.22\textwidth]{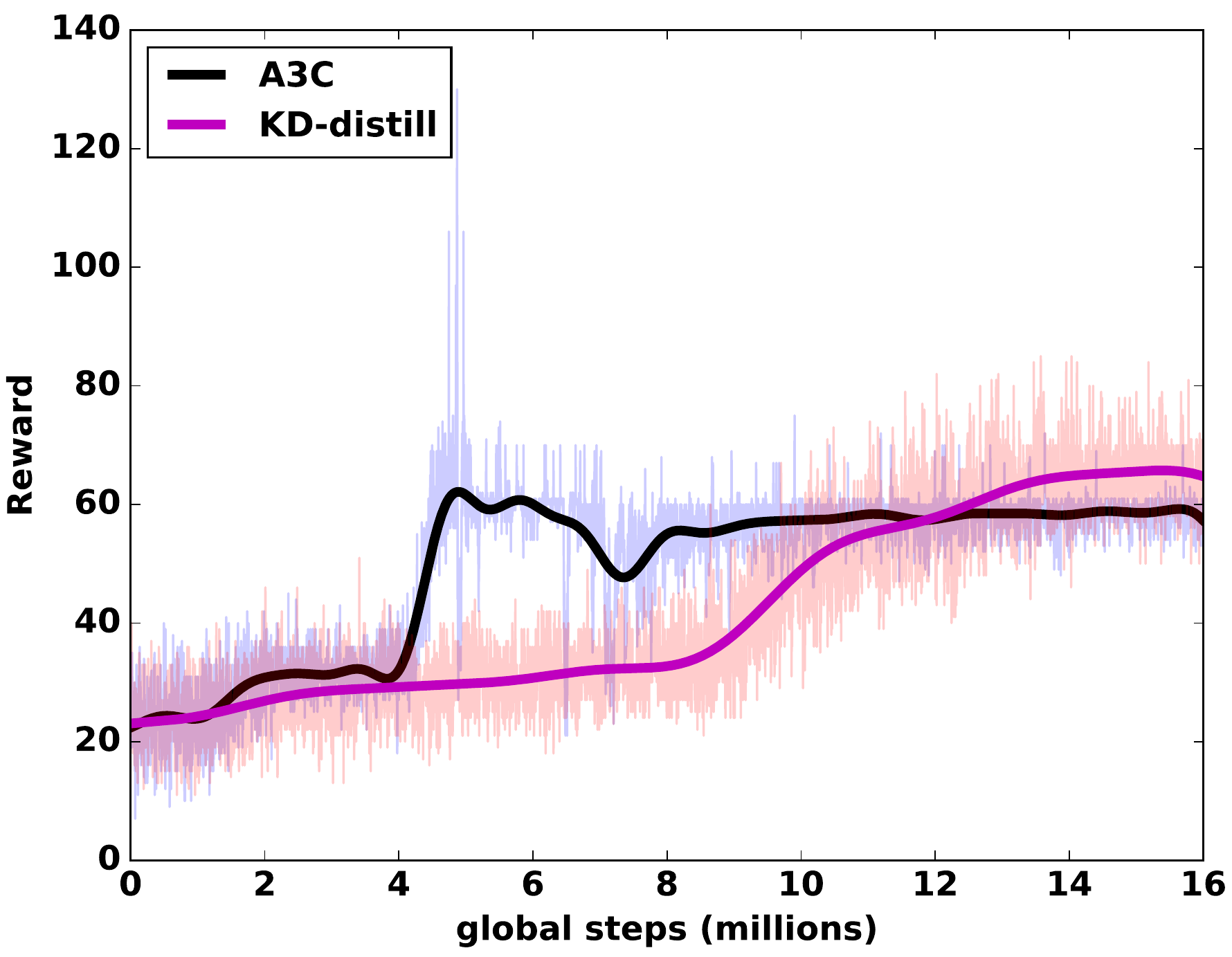} \\
(a) KD with policy layer  & (b) KD with distillation layer   
\end{tabular}
\vspace{-0.15in}
\caption{Performance of online knowledge distillation from a heterogeneous
task. 
(a) distillation from a \textsc{Pong} expert using the policy
layer to train a \textsc{Bowling} student (KD-policy).
(b) distillation from a \textsc{Pong} expert to a
\textsc{Bowling} student using an extra distillation layer (KD-distill).
}
\label{fig:KD_effect}
\vspace{-0.25in}
\end{figure}

\begin{figure}
\centering
\begin{tabular}{c c c}
\includegraphics[width=0.14\textwidth]{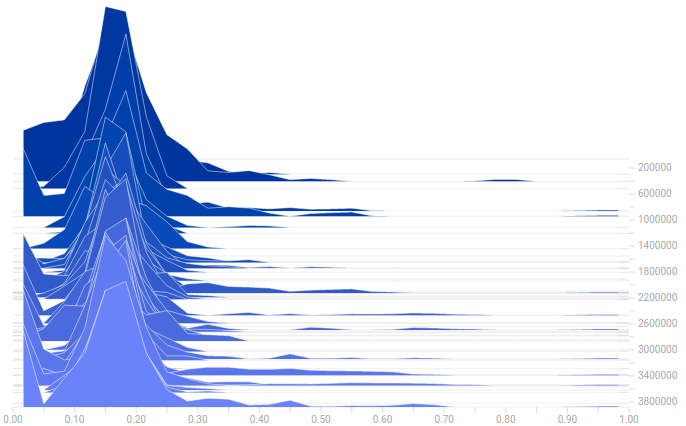} &
\includegraphics[width=0.14\textwidth]{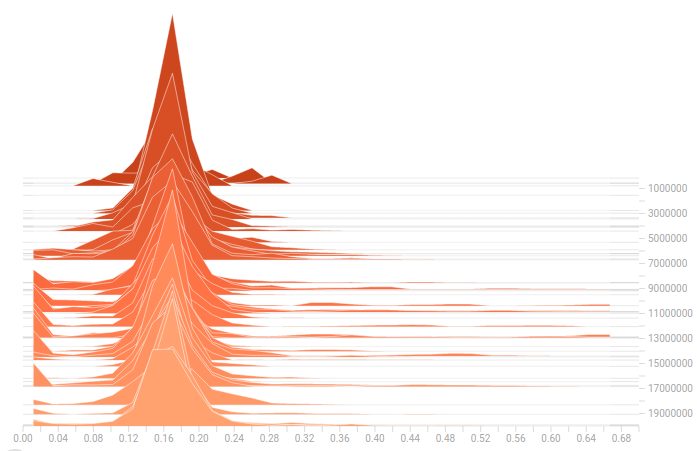} &
\includegraphics[width=0.14\textwidth]{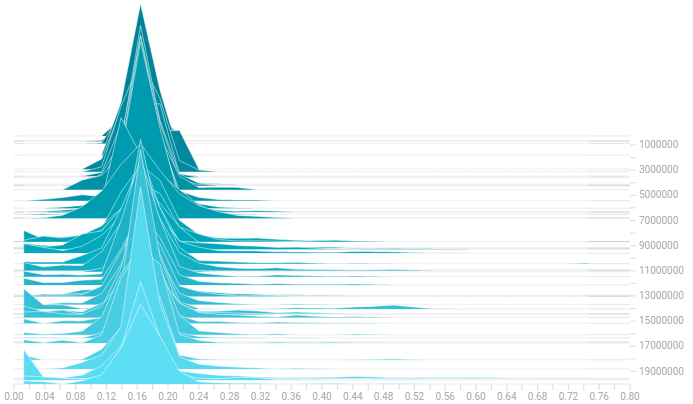} \\
(a) \textsc{Pong}  & (b) \textsc{Bowling}  & (c) aligned \textsc{Pong}
\end{tabular}
\vspace{-0.1in}
\caption{The action probability distributions of a \textsc{Pong} expert, 
a \textsc{Bowling} expert and an aligned \textsc{Pong} expert. }
\label{fig:action_prob}
\vspace{-0.15in}
\end{figure}

In this subsection, we design experiments to illustrate the effectiveness of
CDRL in certificated heterogeneous transfer, with the proposed deep knowledge
distillation.
Given a certificated task \textsc{Pong}, we want to utilize the existing
expertise and apply it to facilitate the training of a new task
\textsc{Bowling}. In the following experiments, we do not tune any 
model-specific parameters such as number of layers, size of filter or network
structure for \textsc{Bowling}. 
We first directly perform transfer learning from \textsc{Pong} to
\textsc{Bowling} by knowledge distillation. Since the two tasks has different action patterns
and action probability distributions, directly knowledge distillation with a
policy layer is not successful, as shown in Figure~\ref{fig:KD_effect}~(a). In
fact, the knowledge distilled from \textsc{Pong} contradicts to the knowledge
learned from \textsc{Bowling}, which leads to the much worse performance than
the baseline. We show in Figure~\ref{fig:action_prob} (a) and (b) that the
action distributions of \textsc{Pong} and \textsc{Bowling} are very different.
To resolve this, we distill the knowledge through an extra distillation layer
as illustrated in Figure~\ref{fig:deepKD}~(b). As such, the knowledge
distilled from the certificated heterogeneous task can be successfully
transferred to the student network with improved performance after the
learning is complete. However, this leads to a much slower convergence than
the baseline as shown in Figure~\ref{fig:KD_effect}~(b), because that it takes
time to learn a good distillation layer to align the knowledge distilled from
\textsc{Pong} to the current learning task. An interesting question is that,
is it possible to have both improved performance and faster convergence? 

\begin{figure*}
\centering
\centering
\begin{tabular}{c c c | c}
\includegraphics[width=0.22\textwidth]{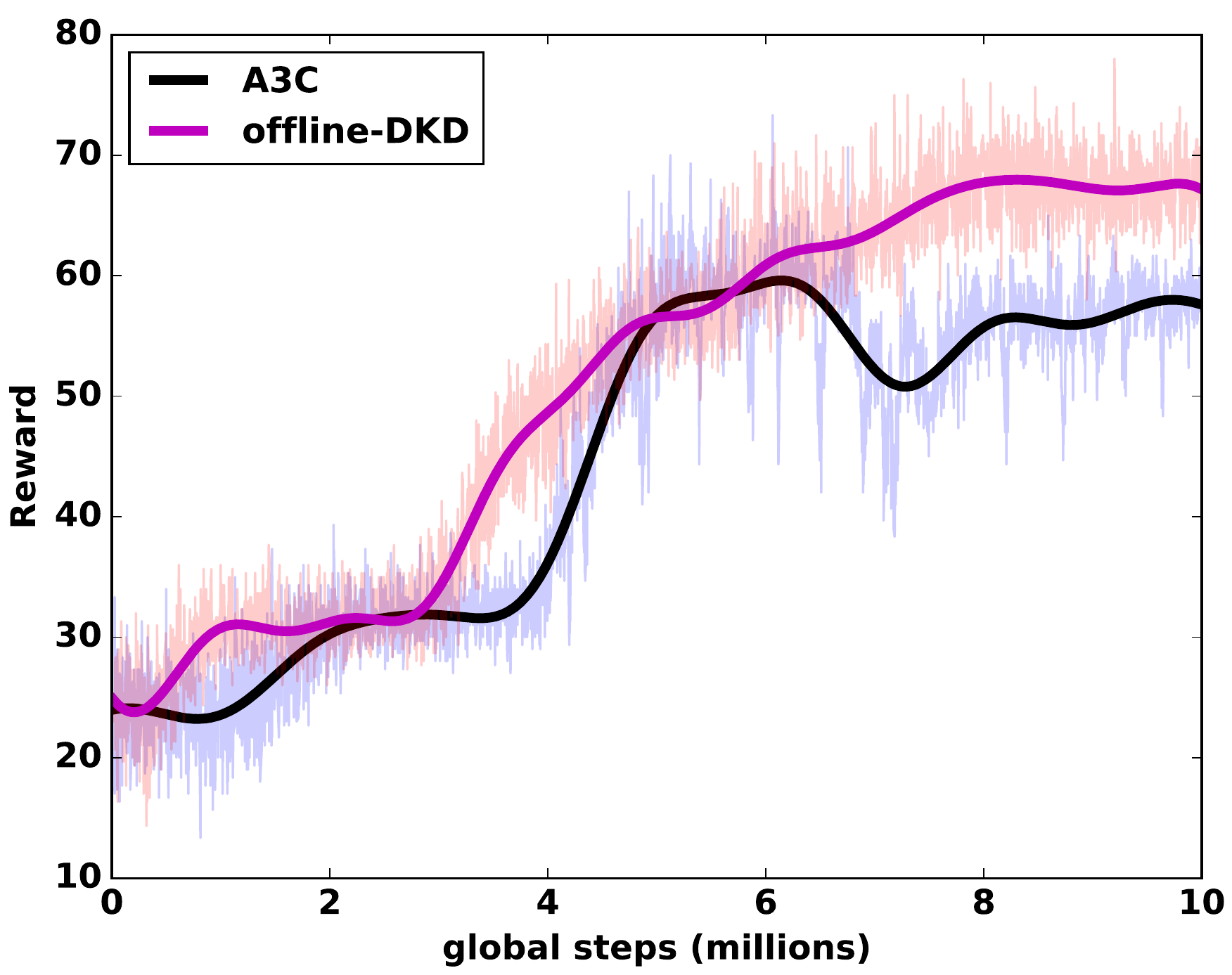} &
\includegraphics[width=0.22\textwidth]{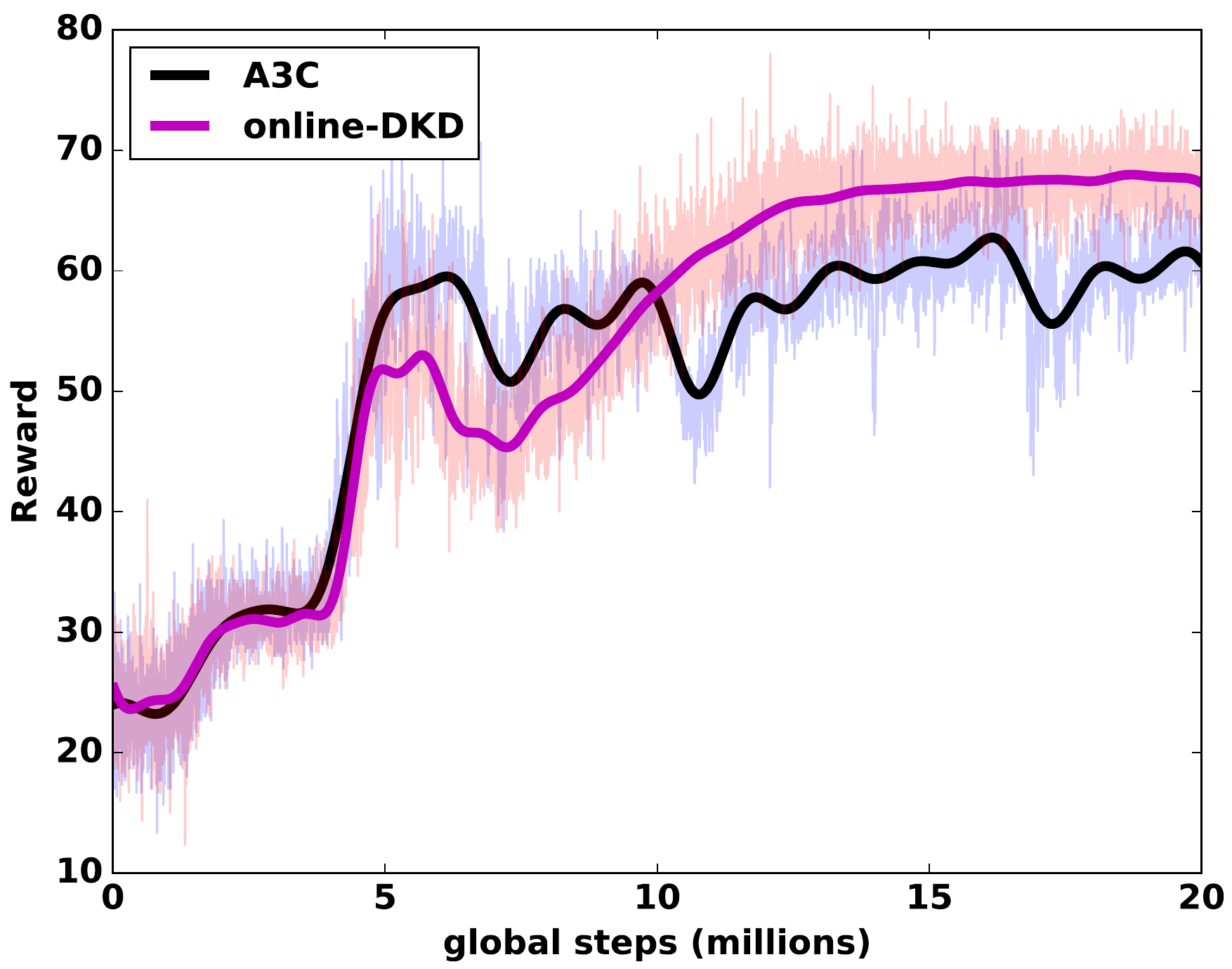} &
\includegraphics[width=0.22\textwidth]{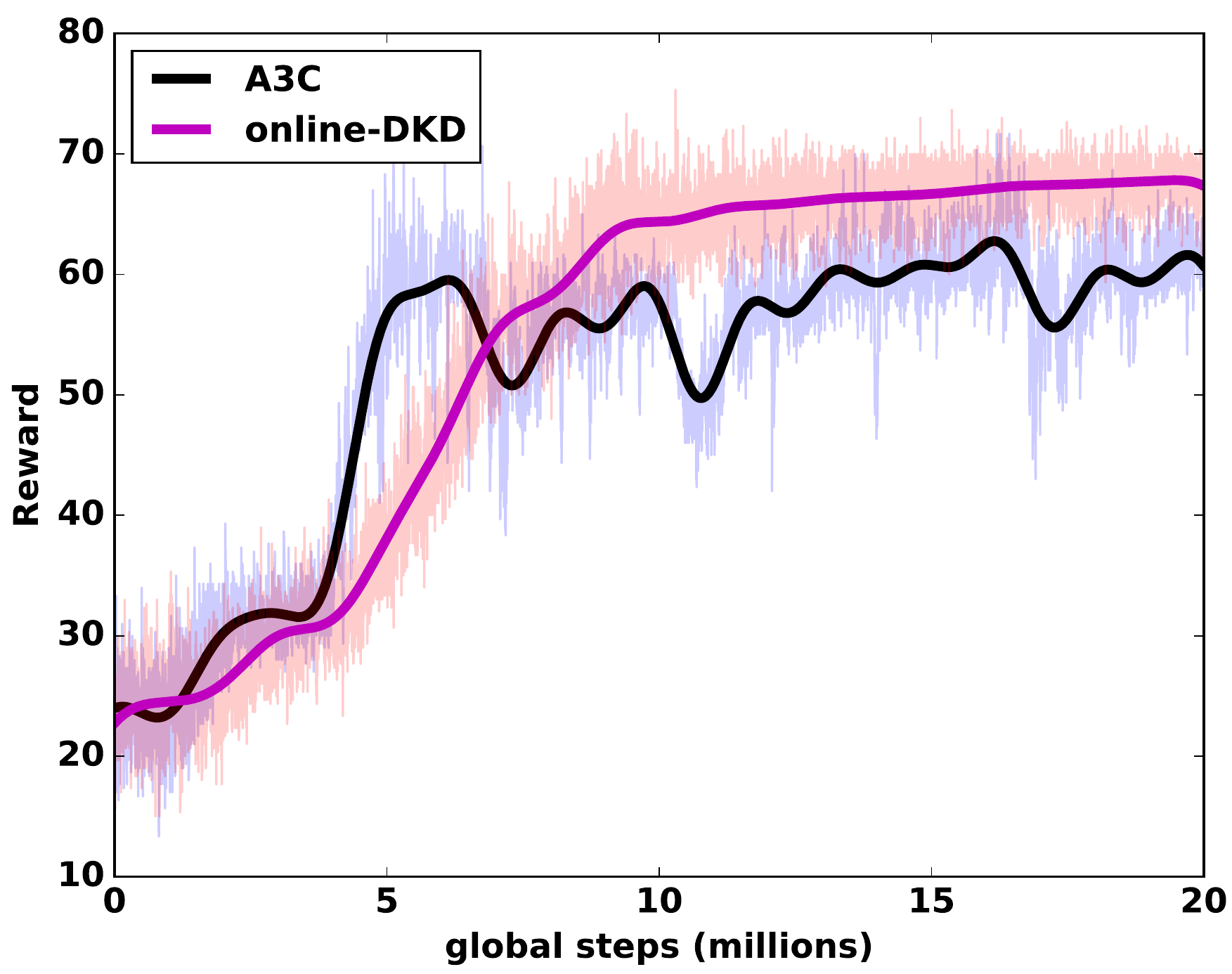} &
\includegraphics[width=0.22\textwidth]{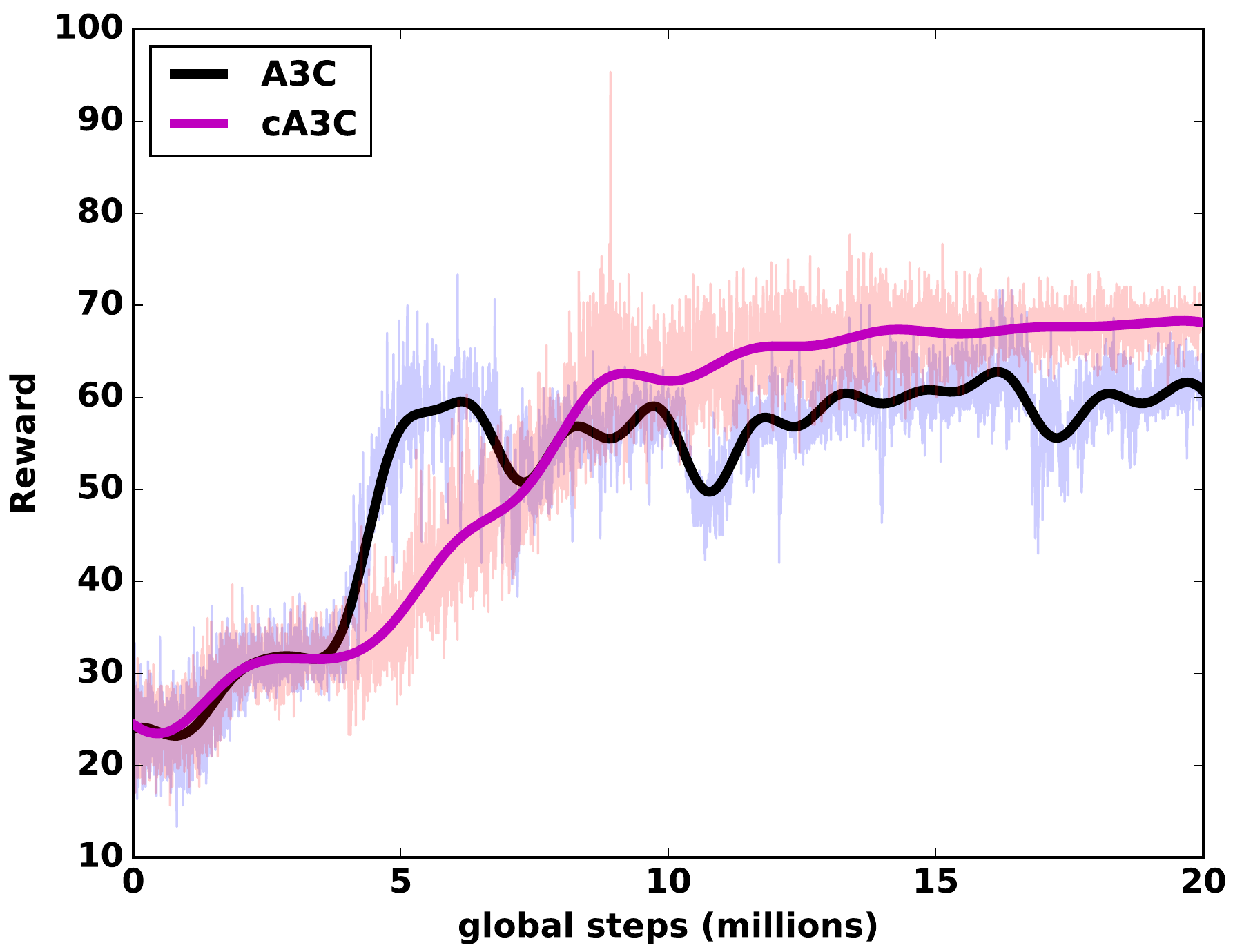} \\
(a) Offline  & (b) Online Strategy 1  & (c) Online Strategy 2 & (d) Collaborative 
\end{tabular}
\vspace{-0.1in}
\caption{Performance of \textbf{offline}, \textbf{online} deep knowledge
distillation, and collaborative learning. Results averaged over 3 runs.}
\label{fig:deep_consolidate}
\vspace{-0.1in}
\end{figure*}

\noindent\textbf{Deep knowledge distillation -- Offline training.}
To handle the heterogeneity between \textsc{Pong} and \textsc{Bowling}, we
first verify the effectiveness of deep knowledge distillation with an offline
training procedure. The offline training is split into two stages. In the
first stage, we train a deep alignment network with four fully connected
layers using the Relu activation function. The training data are logits
generated from an expert \textsc{Pong} network and \textsc{Bowling} network.
The rewards of the networks at convergence are 20 and 60 respectively. 
In stage 2, with the \textsc{Pong} teacher network and trained deep alignment
network, we train a \textsc{Bowling} student network from scratch. The student
network is trained with both GAE interactions with its environment, and the
distillation interactions from the teacher network and the deep alignment
network. The results in 
Figure~\ref{fig:deep_consolidate}~(a)
show that deep knowledge
distillation can transfer knowledge from \textsc{Pong} to \textsc{Bowling}
both efficiently and effectively. 

\noindent\textbf{Deep knowledge distillation -- Online training.}
A more practical setting of CDRL is the online training, where we
simultaneously train deep alignment network and conduct the online deep
knowledge distillation. We use two online training strategies: 1) The training
of deep alignment network starts after 4 million steps, when the student
\textsc{Bowling} network can perform reasonably well, and the knowledge
distillation starts after 6 million steps. 2) The training of deep alignment
network starts after 0.1 million steps, and the knowledge distillation starts
after 1 million steps. Results are shown in Figure~\ref{fig:deep_consolidate}
(b) and (c) respectively. The results show that both strategies reach higher
performance than the baseline. Moreover, the results suggest that we do not
have to wait until the student network reaches a reasonable performance before
we start to train the deep alignment network. This is because the deep
alignment network is train to align two distributions of \textsc{Pong} and
\textsc{Bowling}, instead of transferring the actual knowledge. Recall 
that the action probability distribution of \textsc{Pong} and \textsc{Bowling}
are quite different as shown in Figure~\ref{fig:action_prob} (a) and (b).
After we projecting the logits of \textsc{Pong} using the deep alignment
network, the distribution is very similar to \textsc{Bowling}, as shown in
Figure~\ref{fig:action_prob} (c).

\vspace{-0.05in}
\subsection{Collaborative Deep Reinforcement Learning}
In previous experiments, we assume that there is a well-trained
\textsc{Pong} expert, and we transfer knowledge from the \textsc{Pong} expert to
the \textsc{Bowling} student via deep knowledge distillation.
A more challenging settings that both of
\textsc{Bowling} and \textsc{Pong} are trained from scratch. In this experiment, we 
we show that the CDRL framework can still be effective in this setting. In
this experiment, we train a \textsc{Bowling} network and a \textsc{Pong}
network from scratch using the proposed cA3C algorithm. 
The \textsc{Pong} agents are trained with GAE interactions only, and the
target \textsc{Bowling} receive supervision from both GAE interactions and
distilled knowledge from \textsc{Pong} via a deep alignment network. 
We start to train the deep alignment network after 3 million steps, 
and perform deep knowledge distillation after 4 million steps,
where the \textsc{Pong} agents are still updating from the environment. We
note that in this setting, the teacher network is constantly being updated, as
knowledge is distilled from the teacher until 15 million steps. Results in
Figure~\ref{fig:deep_consolidate}~(d) show that the proposed cA3C is able to
converge to a higher performance than the current state-of-art. The reward of
last one hundred episodes of A3C is $61.48 \pm 1.48$, while cA3C achieves
$68.35 \pm 1.32$, with a significant reward improvement of $11.2\%$.









\section{Conclusion}
In conclusion, we propose a collaborative deep reinforcement learning framework
that can address the knowledge transfer among heterogeneous tasks. 
Under this framework, we propose deep knowledge distillation to adaptively align
the domain of different tasks with the utilization of deep alignement network. 
Furthermore, we develeop an efficient cA3C algorithm and demonstrate its effectiveness 
by extensive evaluation on OpenAI gym. 

\appendix

\section*{Acknowledgments}
This research is supported in part by the Office of Naval Research (ONR) under
grant number N00014-14-1-0631 and National Science Foundation under Grant
IIS-1565596, IIS-1615597. 

\vspace{-0.1in}

\bibliographystyle{ACM-Reference-Format}
\bibliography{ref.bib}  

\end{document}